\RequirePackage[log]{snapshot}
\documentclass[wcp]{jmlr}
\usepackage[utf8]{inputenc}
\usepackage{longtable}% for long tables

\usepackage{booktabs}
\usepackage[load-configurations=version-1]{siunitx} % newer version
\usepackage{paralist}
\DeclareRobustCommand{\inlinelist}[1]{\begin{inparaenum}[(a)] #1 \end{inparaenum}}
\usepackage{rotating}
\usepackage{multirow}
\DeclareMathOperator*{\argmin}{arg\,min}
 
\jmlryear{2014}
\jmlrworkshop{ICML 2014 AutoML Workshop}

\title[Automatic Differentiation of Algorithms for Machine Learning]{Automatic Differentiation of Algorithms\\ for Machine Learning}

\author{\Name{Atılım Güneş Baydin} \Email{atilimgunes.baydin@nuim.ie}\\
\Name{Barak A. Pearlmutter} \Email{barak@cs.nuim.ie}\\\addr Department of Computer Science \& Hamilton Institute\\National University of Ireland Maynooth, Co.\ Kildare, Ireland
}

\begin{document}

\maketitle

\begin{abstract}
Automatic differentiation---the mechanical transformation of numeric computer programs to calculate derivatives efficiently and accurately---dates to the origin of the computer age. Reverse mode automatic differentiation both antedates and generalizes the method of backwards propagation of errors used in machine learning. Despite this, practitioners in a variety of fields, including machine learning, have been little influenced by automatic differentiation, and make scant use of available tools. Here we review the technique of automatic differentiation, describe its two main modes, and explain how it can benefit machine learning practitioners. To reach the widest possible audience our treatment assumes only elementary differential calculus, and does not assume any knowledge of linear algebra.
\end{abstract}

\section{Introduction}

Many methods in machine learning require the evaluation of derivatives. This is particularly evident when one considers that most traditional learning algorithms rely on the computation of gradients and Hessians of an objective function, with examples in artificial neural networks (ANNs), natural language processing, and computer vision \citep{Sra2011}.

Derivatives in computational models are handled by four main methods: \inlinelist{\item working out derivatives manually and coding results into computer; \item numerical differentiation; \item symbolic differentiation using computer algebra; and \item automatic differentiation.}

Machine learning researchers devote considerable effort for the manual derivation of analytical derivatives for a novel model they introduce, subsequently using these in standard optimization procedures such as L-BFGS or stochastic gradient descent. Manual differentiation has the advantage of avoiding approximation errors and instability known to be present in numerical differentiation, but can be prone to error and labor intensive. Symbolic computation methods address weaknesses of both manual and numerical methods, but often result in complex and cryptic expressions plagued with the problem of ``expression swell''.

The fourth technique, automatic differentiation (AD)\footnote{Also called ``algorithmic differentiation'' and less frequently ``computational differentiation''.} works by systematically applying the chain rule of calculus at the elementary operator level. AD allows accurate evaluation of derivatives with only a small constant factor of overhead and ideal asymptotic efficiency. Unlike the need for arranging algorithms into monolithic mathematical expressions for symbolic differentiation, AD can be applied to existing code with minimal change. Owing to this, it is an established tool in applications such as real-parameter optimization \citep{Walther2007}, sensitivity analysis, and probabilistic inference \citep{Neal2011}.

Despite its widespread use in other fields, AD has been underused, if not unknown, by the machine learning community. How relevant AD can be for machine learning tasks is exemplified by the backpropagation method for ANNs, modeling learning as gradient descent in ANN weight space and utilizing the chain rule to propagate error values. The resulting algorithm can be obtained by transforming the network evaluation function through reverse mode AD. Thus, a modest understanding of the mathematics underlying the backpropagation method gives one already sufficient basis to grasp the technique.

Here we review AD from a machine learning perspective and bring up some possible applications in machine learning. It is our hope that the review will be a concise introduction to the technique for machine learning practitioners.

\section{What AD Is Not}
\label{SectionWhatADIsNot}

\begin{figure}
  \centering
  \subfigure{\resizebox{0.8\textwidth}{!}{\includegraphics{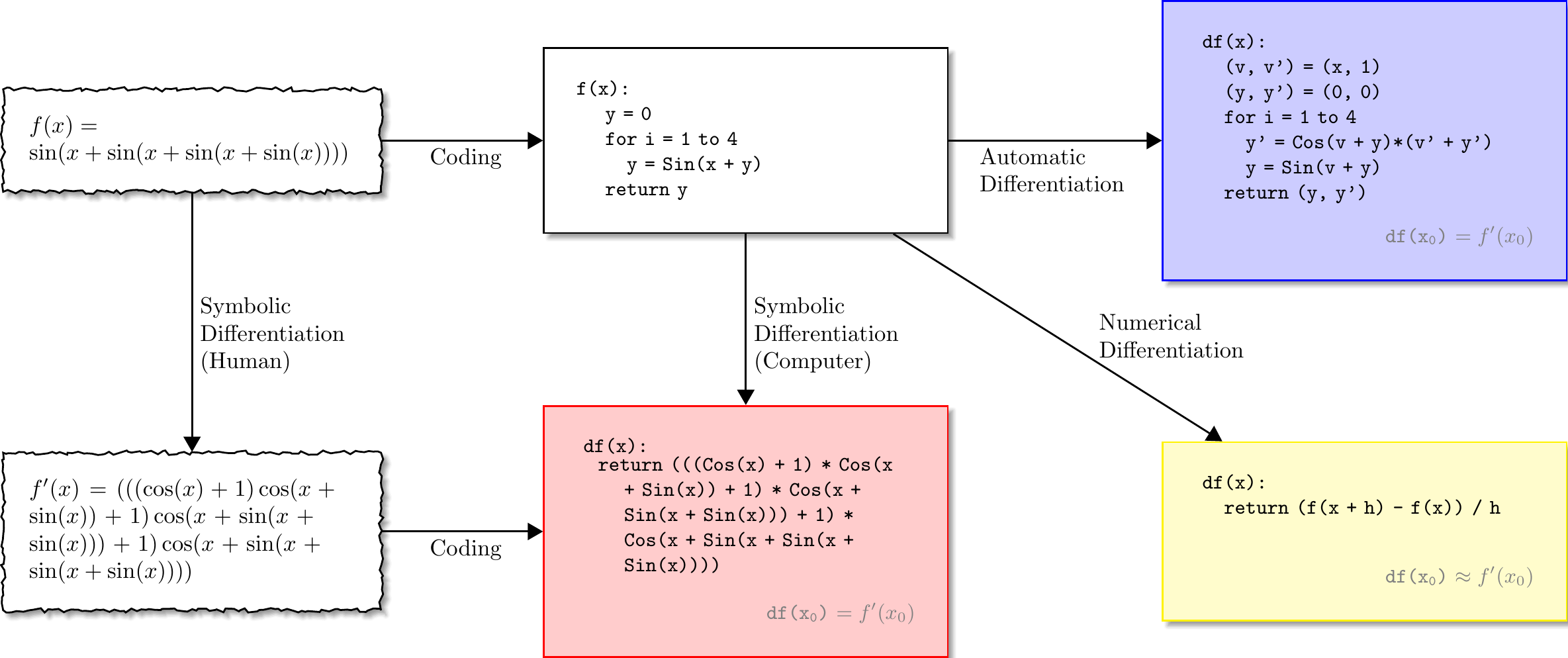}}}\qquad
  \subfigure{\resizebox{0.148\textwidth}{!}{\includegraphics{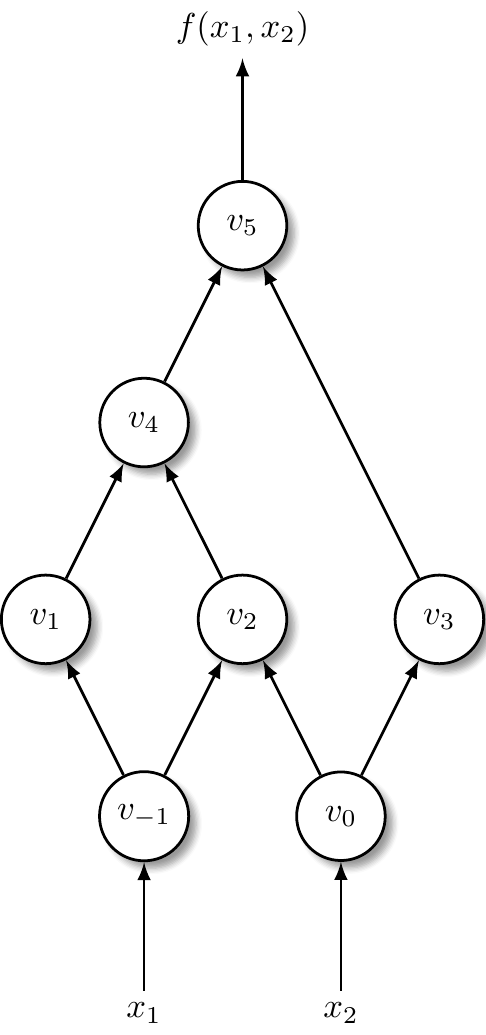}}}
  \caption{(\emph{a}) Differentiation of mathematical expressions and code. Symbolic differentiation (lower center); numerical differentiation (lower right); AD (upper right). (\emph{b}) Computational graph of the example $f(x_1, x_2) = \ln(x_1) + x_1 x_2 - \sin(x_2)$.}
  \label{FigureDifferentiation}
\end{figure}

The term ``automatic differentiation'' has undertones that it is either symbolic or numerical differentiation. The output of AD is indeed numerical derivatives, while the steps in its computation do depend on algebraic manipulation, giving it a two-sided nature partly symbolic and partly numerical. Let us start by stressing how AD is different from, and in some aspects superior to, these two commonly encountered techniques (Figure~\ref{FigureDifferentiation} (a)).

\textbf{\emph{AD is not numerical differentiation.}} Finite difference approximation of derivatives uses the original function evaluated at sample points. In its simplest form, it uses the standard definition $\frac{df(x)}{dx} = \lim_{h \to 0} \frac{f(x + h) - f(x)}{h}$ and approximates the left-hand side by evaluating right-hand side with a small nonzero $h$. This is easy to implement, but inherently prone to truncation and round-off errors. Truncation tends to zero as $h \rightarrow 0$; however, at the same time, round-off increases and becomes dominant. Improvements such as higher-order finite differences or Richardson extrapolation do not completely eliminate approximation errors.

\textbf{\emph{AD is not symbolic differentiation.}} One can generate exact symbolic derivatives through manipulation of expressions via differentiation rules such as $\frac{d}{dx}(u(x) v(x)) = \frac{du(x)}{dx}v(x) + u(x)\frac{dv(x)}{dx}$. This perfectly mechanistic process is realized in computer algebra systems such as Mathematica, Maple, and Maxima. Symbolic results can give insight into the problem and allow analytical solutions of optima (e.g. $\frac{df(x)}{dx}=0$) in which case derivatives are no longer needed. Then again, they are not always efficient for run-time calculations, as expressions can get exponentially larger through differentiation (``expression swell'').

\section{AD Origins}
\label{SectionADOrigins}

For accurate numerical derivatives, it is possible to simplify symbolic calculations by only storing values of intermediate steps in memory. For efficiency, we can interleave, as much as possible, the differentiation and storage steps. This ``interleaving'' idea forms the basis of ``Forward Accumulation Mode AD'': apply symbolic differentiation to each elementary operation, keeping intermediate numerical results, in lockstep with the evaluation of the original function. 

\subsection{Forward Mode}

\begin{table}
  \caption{Forward AD example, with $y = f(x_1, x_2) = \ln(x_1) + x_1 x_2 - \sin(x_2)$ at $(x_1, x_2) = (2, 5)$ and setting $\dot{x}_1 = 1$ to compute $\partial y / \partial x_1$.}
  \label{TableForwardADExample}
  \begin{minipage}[c]{0.48\textwidth}
    \color{black}
    {\scriptsize
    \begin{tabular}{llll}
      \toprule
      \multicolumn{4}{l}{Forward evaluation trace}\\
      \multirow{9}{2mm}{\includegraphics{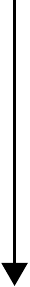}} & $v_{-1}$ & $=x_1$ & $=2$\\
      & $v_0$ & $=x_2$ & $=5$\\
      \cmidrule{2-4}
      & $v_1$ & $=\ln{v_{-1}}$ & $=\ln{2}$\\
      & $v_2$ & $=v_{-1} \times v_0$ & $=2 \times 5$\\
      & $v_3$ & $=\sin{v_0}$ & $=\sin{5}$\\
      & $v_4$ & $=v_1+v_2$ & $=0.6931+10$\\
      & $v_5$ & $=v_4-v_3$ & $=10.6931+0.9589$\\
      \cmidrule{2-4}
      & $y$ & $=v_5$ & $=11.6521$\\
      \bottomrule
    \end{tabular}}
  \end{minipage}
  \begin{minipage}[c]{0.48\textwidth}
    \color{blue}
    {\scriptsize
    \begin{tabular}{llll}
      \toprule
      \multicolumn{4}{l}{Forward derivative trace}\\
      \multirow{9}{2mm}{\includegraphics{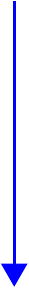}} & $\dot{v}_{-1}$ & $=\dot{x}_1$ & $=1$\\
      & $\dot{v}_0$ & $=\dot{x}_2$ & $=0$\\
      \cmidrule{2-4}
      & $\dot{v}_1$ & $=\dot{v}_{-1}/v_{-1}$ & $=1/2$\\
      & $\dot{v}_2$ & $=\dot{v}_{-1} \times v_0 + v_{-1} \times \dot{v}_0$ & $=1 \times 5 + 2 \times 0$\\
      & $\dot{v}_3$ & $=\cos{v_0}\times \dot{v}_0$ & $=\cos{5}\times 0$\\
      & $\dot{v}_4$ & $=\dot{v}_1+\dot{v}_2$ & $=0.5+5$\\
      & $\dot{v}_5$ & $=\dot{v}_4-\dot{v}_3$ & $=5.5-0$\\
      \cmidrule{2-4}
      & \boldmath$\dot{y}$ & \boldmath$=\dot{v}_5$ & \boldmath$=5.5$\\
      \bottomrule
      \end{tabular}}
  \end{minipage}
\end{table}

All computations are ultimately compositions of a finite set of elementary operations with known derivatives. Combining derivatives of constituent operations through the chain rule gives the derivative of the overall composition. In Table~\ref{TableForwardADExample}, we have the example $f(x_1, x_2) = \ln(x_1) + x_1 x_2 - \sin(x_2)$ represented as an \emph{evaluation trace} of elementary operations---also called a Wengert list. Using the ``three-part'' notation of \citet{Griewank2008}, a trace of $f:\mathbb{R}^n \rightarrow \mathbb{R}^m$ is constructed from \inlinelist{\item $v_{i-n} = x_i, i = 1, \dotsc, n$ input variables, \item $v_i, i = 1, \dotsc, l$ working variables, and \item $y_{m-i} = v_{l-i}, i = m - 1, \dotsc, 0$ output variables.} We can also represent a given trace of operations as a data flow graph, as shown in Figure~\ref{FigureDifferentiation} (b), which makes the dependency relations between intermediate variables explicit.

For computing the derivative with respect to, say $x_1$, we associate with each variable $v_i$ a corresponding $\dot{v}_i = \frac{\partial v_i}{\partial x_1}$. Applying the chain rule to each elementary operation in the forward trace, we generate the derivative trace on the right-hand side. Evaluating variables $v_i$ one by one together with $\dot{v}_i$ gives us the required derivative in the final variable $\dot{v}_5$.

In general, for an $f: \mathbb{R}^n \rightarrow \mathbb{R}^m$ with $n$ independent $x_i$ as inputs and $m$ dependent $y_j$ as outputs, each forward pass of AD is initialized by setting the derivative of only one of inputs $\dot{x}_i = 1$. With given values of $x_i$, a forward run would then compute derivatives of $\dot{y}_j = \frac{\partial y_j}{\partial x_i}, j = 1, \dotsc, m$. Forward mode is ideal for functions $f: \mathbb{R} \rightarrow \mathbb{R}^m$, as all the required derivatives $\frac{\partial y_j}{\partial x}$ can be calculated with one forward pass. Conversely, in the other extreme of $f: \mathbb{R}^n \rightarrow \mathbb{R}$, forward mode would require $n$ forward passes to compute all $\frac{\partial y}{\partial x_i}$. In general, for $f: \mathbb{R}^n \rightarrow \mathbb{R}^m$ where $n \gg m$, reverse AD is faster.

\subsection{Reverse Mode}

Like its familiar cousin backpropagation, reverse AD works by propagating derivatives backward from an output. It does this by supplementing each $v_i$ with an adjoint $\bar{v}_i = \frac{\partial y_j}{\partial v_i}$ representing the sensitivity of output $y_j$ to $v_i$. Derivatives are found in two stages: First, the original function is evaluated \emph{forward}, computing $v_i$ that will be subsequently needed. Second, derivatives are calculated in \emph{reverse} by propagating $\bar{v}_i$ from the output to the inputs. In Table~\ref{TableReverseADExample}, the backward sweep of adjoints on the right-hand side starts with $\frac{\partial y}{\bar{v}_5} = \bar{y} = 1$ and we get both derivatives $\frac{\partial y}{\partial x_1}$ and $\frac{\partial y}{\partial x_2}$ in just one reverse sweep.

\begin{table}
  \caption{Reverse AD example, with $y = f(x_1, x_2) = \ln(x_1) + x_1 x_2 - \sin(x_2)$ at $(x_1, x_2) = (2, 5)$. Setting $\bar{y} = 1$, $\partial y / \partial x_1$ and $\partial y / \partial x_2$ are computed in one reverse sweep.}
  \label{TableReverseADExample}
  \begin{minipage}[c]{0.4\textwidth}
    \color{black}
    {\scriptsize
    \begin{tabular}{p{1mm}p{3mm}p{15mm}p{26mm}}
      \toprule
      \multicolumn{4}{l}{Forward evaluation trace}\\
      \multirow{9}{1mm}{\includegraphics{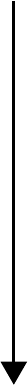}} & $v_{-1}$ & $=x_1$ & $=2$\\
      & $v_0$ & $=x_2$ & $=5$\\
      \cmidrule{2-4}
      & $v_1$ & $=\ln{v_{-1}}$ & $=\ln{2}$\\
      & $v_2$ & $=v_{-1} \times v_0$ & $=2 \times 5$\\
      \\
      & $v_3$ & $=\sin{v_0}$ & $=\sin{5}$\\
      & $v_4$ & $=v_1+v_2$ & $=0.6931+10$\\
      \\
      & $v_5$ & $=v_4-v_3$ & $=10.6931+0.9589$\\
      \\
      \cmidrule{2-4}
      & $y$ & $=v_5$ & $=11.6521$\\
      \bottomrule
    \end{tabular}}
  \end{minipage}
  \begin{minipage}[c]{0.52\textwidth}
    \color{blue}
    {\scriptsize
    \begin{tabular}{p{1mm}p{3mm}p{31mm}p{21mm}p{15mm}@{}}
      \toprule
      \multicolumn{5}{l}{Reverse adjoint trace}\\
      \multirow{9}{1mm}{\includegraphics{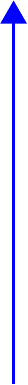}} & \boldmath$\bar{x}_1$ & \boldmath$=\bar{v}_{-1}$ & \boldmath$=5.5$\\
      & \boldmath$\bar{x}_2$ & \boldmath$=\bar{v}_0$ & \boldmath$=1.7163$\\
      \cmidrule{2-5}
      & $\bar{v}_{-1}$ & $=\bar{v}_{-1} + \bar{v}_1(\partial v_1 / \partial v_{-1})$ & $=\bar{v}_{-1} + \bar{v}_1 / v_{-1}$ & $=5.5$\\
      & $\bar{v}_0$ & $=\bar{v}_0 + \bar{v}_2(\partial v_2 / \partial v_0)$ & $=\bar{v}_0 + \bar{v}_2 \times v_{-1}$ & $=1.7163$\\
      & $\bar{v}_{-1}$ & $=\bar{v}_2(\partial v_2 / \partial v_{-1})$ & $=\bar{v}_2 \times v_0$ & $=5$\\
      & $\bar{v}_0$ & $=\bar{v}_3(\partial v_3 / \partial v_0)$ & $=\bar{v}_3 \times \cos{v_0}$ & $=-0.2837$\\
      & $\bar{v}_2$ & $=\bar{v}_4(\partial v_4 / \partial v_2)$ & $=\bar{v}_4 \times 1$ & $=1$\\
      & $\bar{v}_1$ & $=\bar{v}_4(\partial v_4 / \partial v_1)$ & $=\bar{v}_4 \times 1$ & $=1$\\
      & $\bar{v}_3$ & $=\bar{v}_5(\partial v_5 / \partial v_3)$ & $=\bar{v}_5 \times (-1)$ & $=-1$\\
      & $\bar{v}_4$ & $=\bar{v}_5(\partial v_5 / \partial v_4)$ & $=\bar{v}_5 \times 1$ & $=1$\\
      \cmidrule{2-5}
      & $\bar{v}_5$ & $=\bar{y}$ & $=1$\\
      \bottomrule
      \end{tabular}}
  \end{minipage}
\end{table}

An advantage of reverse mode is that it is significantly less costly to evaluate than forward mode for functions with a large number of input variables---at least, in terms of operation count. In the case of $f: \mathbb{R}^n \rightarrow \mathbb{R}$, only one application of reverse mode would be sufficient to compute all partial derivatives $\frac{\partial y}{\partial x_i} = \bar{x}_i$, compared with the $n$ sweeps that forward mode would need. In general, for an $f: \mathbb{R}^n \rightarrow \mathbb{R}^m$, if $time(f)$ is required for evaluating $f$, the time it takes to calculate the $m \times n$ Jacobian by forward AD is $O(n\,time(f))$, whereas the same can be done via reverse AD in $O(m\,time(f))$. That is to say, reverse mode AD performs better when $m \ll n$. On the other hand, Forward AD has only a constant factor overhead in space, while Reverse AD requires storage of intermediate results which increases its space complexity.

\section{Derivatives and Machine Learning}
\label{SectionDerivativesAndMachineLearning}

Machine learning applications where computation of derivatives is necessary can include optimization, regression analysis, ANNs, support vector machines, clustering, and parameter estimation. Let us examine some main uses of derivatives in machine learning and how these can benefit from the use of AD.

Given a function $f: \mathbb{R}^n \rightarrow \mathbb{R}$, classical gradient descent has the goal of finding a (local) minimum $w^* = \argmin_w f(w)$ via updates $\Delta w = -\eta \frac{df}{dw}$, where $\eta$ is the step size. These methods make use of the fact that $f$ decreases steepest if one goes in the direction of the negative gradient. For large $n$, reverse mode AD provides a highly efficient and exact method for gradient calculation, as we have outlined.

More sophisticated quasi-Newton methods, such as the BFGS algorithm and its variant L-BFGS, use both the gradient and the Hessian $H$ of a function. In practice, the full Hessian is not computed but approximated using rank-one updates derived from gradient evaluations. AD can be used here for efficiently computing an exact Hessian-vector product $H v$, via applying forward mode on a gradient found through reverse mode. Thus, $H v$ is computed with $O(n)$ complexity, even though $H$ is a $n \times n$ matrix \citep{Pearlmutter1994}. Hessians arising in large-scale applications are typically sparse. This sparsity, along with symmetry, can be exploited by AD techniques such as elimination on computational graph of the Hessian \citep{Dixon1991} or matrix coloring and compression \citep{Gebremedhin2009}.

Another approach for improving the asymptotic rate of convergence of gradient methods is to use gain adaptation methods such as stochastic meta-descent (SMD), where stochastic sampling is introduced to avoid local minima. An example using SMD with AD is given by \citet{Vishwanathan2006} on conditional random fields (CRF), for the probabilistic segmentation of data.

In ANNs, training is an optimization task with respect to the set of weights, which can in principle be attacked via any method including stochastic gradient descent or BFGS \citep{Apostolopoulou2009}. As we have pointed out, the highly successful backpropagation algorithm is a special case of reverse mode AD and there are instances in literature---albeit few---where ANNs are trained with explicit reference to AD, such as \citet{Eriksson1998} using AD for large-scale feed-forward networks, and \citet{Yang2008} where AD is used to train an ANN-based PID controller.
Beyond backpropagation, the generality of AD opens up new possibilities. An example is given for continuous time recurrent neural networks (CTRNN) by \citet{AlSeyab2008}, where AD is used for training CTRNNs predicting dynamic behavior of nonlinear processes in real time. AD is used to calculate derivatives higher than second order, resulting in significantly reduced network training times as compared with other methods.

In computer vision, first and second order derivatives play an important role in tasks such as edge detection and sharpening \citep{Russ2010}. However, in most applications, these fundamental operations are applied on discrete functions of integer coordinates, approximating those derived on a hypothetical continuous spatial image function. As a consequence, derivatives are approximated using numerical differences.
On the other hand, some computer vision tasks can be formulated as minimization of appropriate energy functionals. This minimization is usually accomplished via calculus of variations and the Euler-Lagrange equation, opening up the possibility of taking advantage of AD. In this area, the first study introducing AD to computer vision is given by \citet{Pock2007} which considers denoising, segmentation, and information recovery from stereoscopic image pairs and notes the benefit of AD in isolating sparsity patterns in large Jacobian and Hessian matrices. \citet{Grabner2008} use reverse AD for GPU-accelerated medical 2D/3D registration, a task concerning the alignment of data from different sources such as X-ray images or computed tomography. A six-fold increase in speed (compared with numerical differentiation using center difference) is reported.

Nested applications of AD would facilitate compositional approaches to machine learning tasks, where one can, for example, perform gradient optimization on a system of many components that can in turn be internally using other derivatives or performing optimization \citep{SiskindPearlmutter2008a, Radul-etal-2012a}. This capability is relevant to, e.g., hyperparameter optimization, where using gradient methods on model selection criteria has been proposed as an alternative to the established grid search and randomized search methods. Examples include the application to linear regression and time-series prediction \citep{Bengio2000} and support vector machines \citep{Chapelle2002}.

It is important to note that AD is applicable to not only mathematical expressions in classical sense, but also algorithms of arbitrary structure, including those with control flow statements (Figure~\ref{FigureDifferentiation}). Computations involving if-statements, loops, and procedure calls are in the end evaluated as straight-line traces of elementary operations---i.e. conditionals turned into actual paths taken, loops unrolled, and procedure calls inlined. In contrast, symbolic methods cannot be applied to such algorithms without significant manual effort.

A concerted effort to generalize AD to make it suitable for a wider range of machine learning tasks has been undertaken \citep{Pearlmutter-Siskind-2008}. The resulting AD-enabled research prototype compilers generate very efficient code \citep[][also DVL \url{https://github.com/axch/dysvunctional-language/}]{Siskind-Pearlmutter-2008-TR1}, but these technologies are not yet available in production-quality systems.

\section{Implementations}

In practice, AD is used via feeding an existing algorithm into a tool, which augments it with the corresponding extra code to compute derivatives. This can be implemented through calls to a library; as a source transformation where a given code is automatically modified; or through operator overloading, which makes the process transparent to the user. Implementations exist for most programming languages\footnote{The website \url{http://www.autodiff.org/} maintains a list of implementations.} and a taxonomy of tools is given by \citet{Bischof2008}.

\section{Conclusions}
\label{SectionConclusions}

The ubiquity of differentiation in machine learning renders AD a highly capable tool for the field. Needless to say, there are occasions where we are interested in obtaining more than just the numerical values for derivatives. Symbolic methods can be useful for analysis and gaining insight into the problem domain. However, for any non-trivial function of more than a handful of variables, analytic expressions for gradients or Hessians increase rapidly in complexity to render any interpretation unlikely.

Combining the expressive power of AD operators and functional programming would allow very concise implementations for a range of machine learning applications, which we intend to discuss in an upcoming article.

\acks{This work was supported in part by Science Foundation Ireland grant 09/IN.1/I2637.}

\vspace{-1cm}
\end{document}